\documentclass{article}

\PassOptionsToPackage{numbers,sort&compress}{natbib}

\usepackage[preprint]{neurips_2026}

\usepackage[utf8]{inputenc}
\usepackage[T1]{fontenc}
\usepackage{hyperref}
\usepackage{url}
\usepackage{booktabs}
\usepackage{amsfonts}
\usepackage{amsmath}
\usepackage{nicefrac}
\usepackage{microtype}
\usepackage{xcolor}
\usepackage{graphicx}
\usepackage{caption}
\usepackage{enumitem}
\usepackage{array}

\newcommand{\cfd}[1]{{\fontencoding{OT1}\fontfamily{cmtt}\selectfont #1}}


\hypersetup{colorlinks=true, linkcolor=black, citecolor=black, urlcolor=blue!50!black,
  pdftitle={AegisBench: Auditing Aerial Search-and-Rescue Person Detection Under the Conditions It Is Deployed In},
  pdfauthor={Amogh Vinaykumar, Prem Kamasani}}

\newcommand{\coderepo}{\url{https://github.com/Altis-2026/SentinelBench}}

\title{AegisBench: Auditing Aerial\\
Search-and-Rescue Person Detection\\
Under the Conditions It Is Deployed In}

\author{%
  Amogh Vinaykumar \\
  Flower Mound High School \\
  \cfd{amogh.vinaykumar@gmail.com} \\
  \And
  Prem Kamasani \\
  Flower Mound High School \\
  \cfd{premkamasani@gmail.com} \\
}

\begin{document}

\maketitle

\begin{abstract}
Drone-based person detection is being adopted by emergency-management
agencies to find missing people faster than ground teams can. The
benchmarks behind that adoption were collected in calm, clear daylight,
while the missions that trigger deployment happen in floods, wildfires,
storms, and at night. AegisBench measures that gap. It applies nine
physically motivated corruptions from four disaster families, each at three
severities monotone in a declared image statistic or physical quantity, to
two aerial search-and-rescue datasets, HERIDAL and SARD. Three
architecturally distinct detectors (Faster R-CNN, YOLOv11, RT-DETR) are
scored across 168 conditions by one shared evaluator, at confidence
thresholds frozen on clean validation data as a fielded system would run.
Robustness is highly uneven. Low light is the extreme case: a 4.5-stop
exposure loss drives recall to exactly zero for all three detectors on both
datasets, a result that holds under threshold-free mAP and across three
training seeds, and is mirrored on 300 real night drone images, where 2 of
759 people are found. Surviving detections largely stay on target, so the
failure is silent: an empty frame looks the same as one whose occupants the
system cannot see. Fine-tuning on milder low-light levels recovers 0.06 to
0.19 recall at the held-out severity, at an architecture-dependent cost on
other conditions, and does not transfer to real night imagery. Spatial
corruptions rank differently on the two datasets, which we cannot attribute
to altitude alone because their parameters scale with frame size rather
than person size. We release the corruption engine, evaluation pipeline,
and all per-condition results as an audit instrument for anyone evaluating
these systems.
\end{abstract}

\section{Introduction}
\label{sec:intro}

Automated person detection in drone imagery is moving out of the
literature and into public emergency management. Agencies fly aerial
search patterns and run detectors over the footage, letting one operator
review coverage that would take a ground team days to walk. The appeal is
straightforward: time is the dominant factor in survival, and incident
data show that most missing subjects are found within the first day, with
the odds of a good outcome falling as the search
lengthens~\citep{koester2008lost}.

The evidence behind that adoption has a gap in it. Published results on
the two standard benchmarks for this task, HERIDAL~\citep{bozic2019heridal}
and SARD~\citep{sambolek2021sard}, report strong detection performance, and
both datasets were collected in calm, clear, well-lit conditions. Real
missions are launched because something has gone wrong: after a flood,
during a wildfire, in a storm, or at night, when visibility is worst and
the need for automated coverage is greatest. No existing benchmark
systematically measures how these detectors behave under the visual
conditions of the disasters that trigger their deployment. An agency
comparing systems and a vendor reporting accuracy are both working from
numbers collected in conditions the system will rarely operate in.

This is a familiar pattern in applied machine learning: a model performs
well on the distribution that justified its adoption and worse on the one
it is deployed against~\citep{koh2021wilds}. What makes the aerial search
case unusually tractable is that the deployment conditions are physically
characterizable. Haze, smoke, rain, muddy water, and darkness have
documented optical models, so the missing evaluation can be built rather
than only argued for.

AegisBench builds it. Nine physically motivated corruptions, spanning
four disaster families (flood, wildfire, storm, and earthquake), are
applied to the HERIDAL and SARD test sets, and three architecturally
distinct detectors are evaluated across all resulting conditions. Three
design choices make the result usable as an audit rather than only as a
research finding. First, each corruption is built on a documented optical
model of a disaster condition, such as the Koschmieder scattering model
for smoke and dust, so a result maps onto a named condition an operator
can recognize. Second, each corruption's three severities are calibrated
against a declared, machine-checked image statistic rather than chosen by
eye, so severity claims are falsifiable. Third, every detector's
confidence threshold is selected once on clean validation data and frozen
across all corrupted conditions, matching how a system is fielded rather
than a best case that is unavailable mid-mission.

\begin{figure}[t]
\centering
\includegraphics[width=\linewidth]{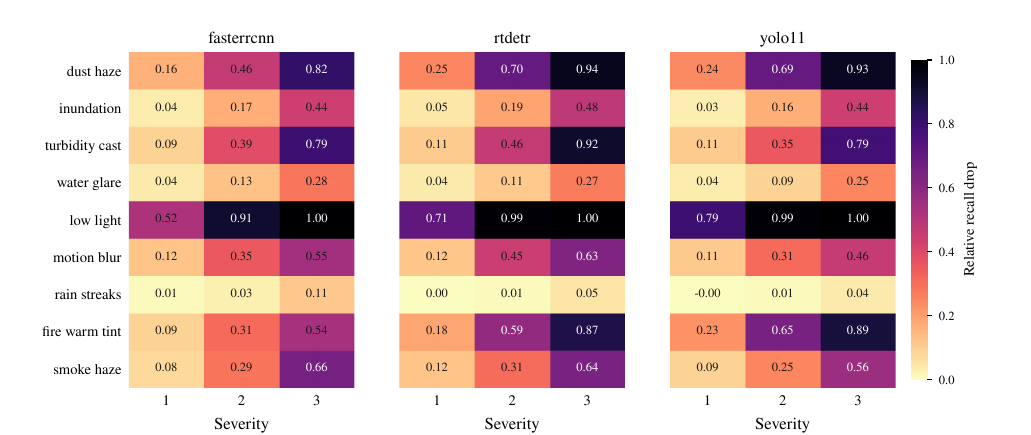}
\caption{Relative recall drop, $(\text{clean}-\text{corrupted})/\text{clean}$,
for each detector on SARD at every corruption and severity, rows grouped by
disaster family; darker is worse. The same three detectors that lose at
most 11\% of their recall under rain lose all of it under low light.}
\label{fig:headline}
\end{figure}

Robustness turns out to be highly uneven (Figure~\ref{fig:headline}).
Low light is the extreme case: it drives recall to exactly zero on both
datasets for all three architectures, a result that holds independently
of the confidence threshold and across training seeds, and that has a
close counterpart on real night drone imagery. The detections a model
does retain largely stay on the person, so the dominant failure is a
detector ceasing to find someone rather than misplacing the box. That
makes it an oversight problem as well as an accuracy problem: a misplaced
box is a visible error an operator can catch, while a frame with no
detections is indistinguishable from a correctly searched empty one.

\paragraph{Contributions.}
\begin{itemize}[leftmargin=1.4em,itemsep=2pt,topsep=2pt]
\item \textbf{A disaster-grounded corruption benchmark.} Nine corruptions
  across four disaster families, each at three severities monotone in a
  declared physical or image statistic, deterministically seeded so the
  corrupted test sets regenerate bit-identically from the public source
  data (Section~\ref{sec:design}).
\item \textbf{A deployment-realistic audit protocol and its results.}
  Frozen operating points, one shared evaluator, bootstrap intervals, and
  a localization-stability axis, applied to three detectors across 168
  conditions. Every per-condition number is released
  (Sections~\ref{sec:protocol}--\ref{sec:results},
  Appendix~\ref{app:full}).
\item \textbf{A cross-architecture low-light collapse}, checked against
  the obvious alternative explanations (operating point, training seed,
  synthetic optics) and against a targeted augmentation, which moves the
  synthetic floor at an architecture-dependent cost elsewhere
  (Section~\ref{sec:lowlight}).
\item \textbf{Evidence that the failure is silent}, and a concrete
  condition-stratified reporting practice for systems used in disaster
  response (Sections~\ref{sec:locstab} and~\ref{sec:deployment}).
\item \textbf{A low-cost, reusable audit instrument.} The corruption
  engine, evaluator and all results are public under an MIT license, run
  on a single 8~GB GPU, and accept any detector behind a common interface,
  so agencies and other groups can audit their own systems under the same
  168 conditions.
\end{itemize}

\section{Related Work}
\label{sec:related}

\paragraph{Aerial search-and-rescue person detection.} This task is most
often benchmarked on HERIDAL~\citep{bozic2019heridal}, high-altitude
imagery of wilderness terrain, and SARD~\citep{sambolek2021sard},
lower-altitude drone video of staged rescue scenarios. Both report strong
accuracy from CNN detectors trained and evaluated under calm, clear
conditions~\citep{bachir2024yolov5sar}. Adjacent benchmarks extend the
small-person-in-large-scene setting to other
domains~\citep{yu2020tinyperson,varga2022seadronessee}. The closest prior
robustness evaluation on this data is the corrupted supplement released
with SARD itself~\citep{sambolek2021sard}, which adds fog, snow, ice, and
motion-blur variants: generic weather effects without a calibrated
severity structure, reported as a single check rather than a systematic
sweep. Outside that supplement, evaluation in this literature is
conducted under clear conditions.

\paragraph{Corruption-robustness benchmarking.} ImageNet-C established
the practice of evaluating models under a fixed suite of synthetic
corruptions at graded severities~\citep{hendrycks2019imagenetc}. Michaelis
et al.~\citep{michaelis2019benchmarking} carried it into object detection
for driving scenes and remain the closest methodological antecedent, and
3D Common Corruptions~\citep{kar20223dcc} showed that corruptions which
respect scene geometry expose failures that flat image filters miss. Our
domain, our disaster-specific and statistically calibrated corruptions,
and our localization-stability axis all differ. In remote sensing, Mei et
al.~\citep{mei2024robustness} benchmark the natural and adversarial
robustness of many classifiers and detectors on overhead imagery, using
generic perturbations rather than condition-specific physical models.
Robustness benchmarking for drone imagery has become active only recently:
UAV-C~\citep{liu2024uavc} benchmarks UAV \emph{tracking} against generic
corruptions, HazyDet~\citep{feng2024hazydet} pairs synthetic haze with real
foggy drone photographs for general object detection, and
WRRT-DETR~\citep{liu2025wrrtdetr} builds a weather-robust detector around a
maritime dataset. Each targets a different task, a narrower corruption
scope, or a different domain than the sweep presented here, and most pair
the benchmark with a proposed architecture. We propose no detector, so the
measurement stands on its own as an audit instrument.

\paragraph{Adverse weather and low-light vision.} Our corruption models
draw on the Koschmieder scattering model as formalized by Narasimhan and
Nayar~\citep{narasimhan2002vision}, the linear-domain treatment of low
light of Chen et al.~\citep{chen2018dark}, and the rain-streak model of
Garg and Nayar~\citep{garg2007rain}. Sakaridis et
al.~\citep{sakaridis2018foggy} apply synthetic fog to real driving
imagery, the closest precedent for corrupting a real dataset with a
physical model, and later contribute real adverse-condition
data~\citep{sakaridis2021acdc}. Dehazing~\citep{he2011dark} and low-light
datasets and enhancement~\citep{loh2019exdark} aim to restore images or
train for darkness rather than to measure how a fielded detector degrades.
Small persons in HERIDAL-scale frames motivate the tiled inference we
adopt~\citep{akyon2022sahi}.

\paragraph{Documentation for deployed systems.} Our framing follows the
model- and dataset-documentation literature, which argues that systems
entering consequential use need standardized statements of intended use
and of the conditions under which performance was
measured~\citep{gebru2021datasheets,mitchell2019modelcards}. Model cards
in particular call for results disaggregated by relevant conditions. For
this task the relevant conditions are environmental, and the disaggregated
number has never been measured, so there has been nothing to put in the
card. We supply the measurement and release the instrument that produced
it.

\section{Benchmark Design}
\label{sec:design}

\begin{figure}[t]
\centering
\includegraphics[width=0.92\linewidth]{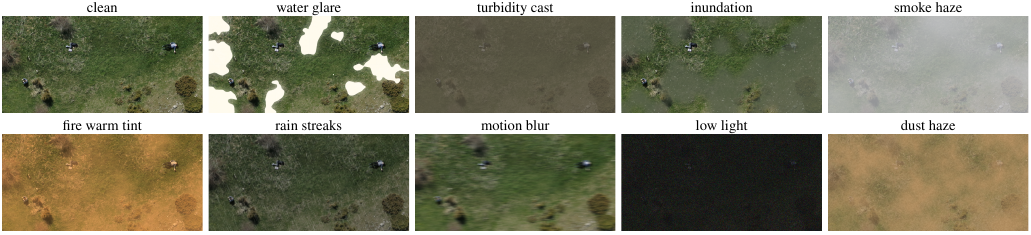}
\caption{The nine corruptions at severity 3 on a cropped HERIDAL frame.
Each models a documented optical signature of a disaster environment
rather than a generic distortion. Severity is calibrated per corruption
against a declared image statistic (Table~\ref{tab:taxonomy}).}
\label{fig:taxonomy}
\end{figure}

Nine corruptions are organized into four disaster families: flood
(\cfd{water\_glare}, \cfd{turbidity\_cast}, \cfd{inundation}), wildfire
(\cfd{smoke\_haze}, \cfd{fire\_warm\_tint}), storm (\cfd{rain\_streaks},
\cfd{motion\_blur}, \cfd{low\_light}), and earthquake and post-disaster
(\cfd{dust\_haze}). Each models a documented optical signature of its
disaster family rather than a generic image-processing filter
(Figure~\ref{fig:taxonomy}, Table~\ref{tab:taxonomy}). Grouping by family
lets a result read as ``this system loses $X$\% of detections in wildfire
smoke,'' which is the form an operational decision needs. Earthquakes have
no optical signature of their own; what degrades post-collapse imagery is
suspended dust, so the family is represented by one physically distinct
corruption rather than padded with relabelled copies of others.

\begin{table}[t]
\centering\small
\caption{The nine corruptions, their disaster family, the optical
mechanism each models, and the statistic defining its severity ladder.
Arrows give the direction the statistic must move as severity increases.
Values are means over a fixed random 20-image subsample of the HERIDAL
test split; monotonicity is verified automatically on these means.
$\dagger$~\cfd{rain\_streaks} has no closed-form statistic isolating
streak density, so this row reports the generation parameter rather than a
measured image statistic (Section~\ref{sec:calibration}).}
\label{tab:taxonomy}
\setlength{\tabcolsep}{3pt}\footnotesize
\begin{tabular}{llllrrr}
\toprule
Corruption & Family & Optical mechanism & Statistic & s1 & s2 & s3 \\
\midrule
\cfd{water\_glare}     & Flood      & Specular sun-glint saturation         & saturated frac.\ $\uparrow$      & 0.0116 & 0.0182 & 0.1895 \\
\cfd{turbidity\_cast}  & Flood      & Sediment-laden water, contrast loss   & RMS contrast $\downarrow$        & 0.1098 & 0.0701 & 0.0378 \\
\cfd{inundation}       & Flood      & Semi-transparent standing water       & occluded frac.\ $\uparrow$       & 0.0830 & 0.2235 & 0.3921 \\
\midrule
\cfd{smoke\_haze}      & Wildfire   & Koschmieder, neutral-gray airlight    & RMS contrast $\downarrow$        & 0.1170 & 0.0776 & 0.0475 \\
\cfd{fire\_warm\_tint} & Wildfire   & Low-CCT firelight, warm airlight haze & red/blue ratio $\uparrow$        & 1.6743 & 2.0721 & 2.4516 \\
\midrule
\cfd{rain\_streaks}    & Storm      & Directional streaks over a rain veil  & streaks/MP $\uparrow$~$\dagger$  & 120    & 350    & 800    \\
\cfd{motion\_blur}     & Storm      & Wind-induced platform shake           & edge strength $\downarrow$       & 0.0764 & 0.0572 & 0.0472 \\
\cfd{low\_light}       & Storm      & Linear-domain photon scaling, noise   & mean luminance $\downarrow$      & 0.2355 & 0.1533 & 0.1175 \\
\midrule
\cfd{dust\_haze}       & Earthquake & Koschmieder, brown mineral airlight   & RMS contrast $\downarrow$        & 0.1094 & 0.0720 & 0.0489 \\
\bottomrule
\end{tabular}
\end{table}

\paragraph{Optical models.} The three haze-family corruptions
(\cfd{smoke\_haze}, \cfd{fire\_warm\_tint}, \cfd{dust\_haze}) use the
Koschmieder atmospheric scattering
model~\citep{koschmieder1924theorie,narasimhan2002vision},
$I = J t + A(1-t)$, with a spatially varying transmission field $t$. They
differ in the chromaticity of the airlight $A$ (neutral gray for
combustion smoke, brown for mineral dust, warm for firelit haze), in the
spatial frequency of the transmission field, and, for dust, in coarse
near-lens grain; the smoke and dust chromaticity difference is enforced by
a unit test on the red-to-blue ratio. Sediment-laden water is not a
scattering process, so the flood family is modeled separately, as a blend
toward mud chromaticity with contrast and saturation loss
(\cfd{turbidity\_cast}), a semi-transparent standing-water layer
(\cfd{inundation}), and specular sun-glint with bloom (\cfd{water\_glare}).
\cfd{low\_light} is modeled in the linear photometric
domain~\citep{chen2018dark}: decode with gamma 2.2, scale photon count,
apply a twilight blue shift, add signal-dependent shot noise and read
noise whose relative magnitude grows as light falls, and re-encode.
\cfd{rain\_streaks} follows the additive directional-streak model of Garg
and Nayar~\citep{garg2007rain} over a blurred, darkened rain veil, and
\cfd{motion\_blur} applies a linear kernel in one random direction per
image, modeling wind-induced platform shake. Appendix~\ref{app:models}
lists every parameter.

\subsection{Calibration and severity}
\label{sec:calibration}

Each corruption declares a measurable image statistic whose value must
move monotonically across severities one through three
(Table~\ref{tab:taxonomy}). Calibration is checked on a fixed random subset
of 20 HERIDAL test images, held constant across all corruptions and
severities so that every ladder is measured on identical scene content. We
measure the declared statistic on every image in the subset and require
its mean to move in the declared direction; the same check runs on
synthetic imagery in the unit tests. This guarantees the subset mean, not
every image: the corruptions are stochastic, so a single image can invert
without the ladder being ill-formed.

Eight of nine corruptions pass this check directly. \cfd{rain\_streaks}
does not have a usable statistic: streak density has no closed-form image
measure, and the substitute edge-strength proxy is confounded, since added
streaks raise edge strength while the veil blur suppresses it. Its
severity is therefore defined by generation parameters, all increasing by
construction, and verified by visual audit; recall degrades in the
expected direction on both datasets regardless (Appendix~\ref{app:full}).

The statistic makes each severity ladder falsifiable \emph{within} a
corruption. It is not a cross-corruption difficulty normalizer and we do
not use it as one: \cfd{smoke\_haze} and \cfd{dust\_haze} reach nearly
identical RMS contrast at every severity (0.0475 versus 0.0489 at
severity 3) and similar optical depth, yet YOLOv11 on SARD keeps
0.391 recall under one and 0.058 under the other. Detection difficulty is
not reducible to global contrast loss.

\paragraph{What the severity levels are, physically.} Where the model
allows, the ladders are stated in physical units. For \cfd{low\_light},
the three severities apply linear photon gains of 0.33, 0.13 and 0.045,
an exposure loss of 1.60, 2.94 and 4.47 stops; severity 3 is a dim dusk
exposure, not darkness. For the haze family, the mean transmission gives
a slant-path optical depth $\tau = -\ln \bar{t}$ of 0.36, 0.80 and 1.39
for \cfd{smoke\_haze} and 0.43, 0.87 and 1.43 for \cfd{dust\_haze}. The
levels were chosen to span mild to severe, not fitted to measurements from
real incidents, so we claim no correspondence between a severity and a
specific event. What the ladders guarantee is monotonicity in a stated
physical quantity, so a reader can place any result against their own
operating conditions.

\paragraph{Scale, alignment, and determinism.} Pixel-unit parameters
(blur kernel length, streak length, bloom width, ripple amplitude) are
specified per 1000 pixels of the longer image side and scaled at runtime,
so a severity is constant relative to the frame. It is not constant
relative to the people in it. At severity 3 the motion-blur kernel is 100
pixels long on a $4000\times3000$ HERIDAL frame and 16 pixels on a
$640\times640$ SARD frame, while most persons in both test sets fall in
the same COCO medium size band~\citep{lin2014coco}, with box areas
between $32^2$ and $96^2$ pixels. This matters for the cross-dataset comparison in
Section~\ref{sec:crossdataset}. Corruptions are appearance-only, so the
clean ground-truth boxes stay exactly valid; the one exception,
\cfd{inundation}'s refraction ripple, is bounded well below person scale
(at most about 1.4 pixels at severity 3 on a 4000-pixel frame). Every
stochastic element is seeded by a SHA-256 hash of
$(\text{image\_id}, \text{corruption}, \text{severity},
\text{global\_seed})$, independent of platform and Python hash
randomization, so the corrupted test sets regenerate bit-identically
anywhere. Nothing derived from HERIDAL or SARD needs to be redistributed
for the benchmark to be reproduced.

\section{Evaluation Protocol}
\label{sec:protocol}

Each choice below closes off a specific way a corruption benchmark can
produce a misleading conclusion, and each corresponds to a way a
vendor-reported robustness number can flatter a system that would not
behave that way in the field.

\paragraph{Frozen operating points.} A detector needs a confidence
threshold to turn scores into detections. We select it once per model per
dataset, by maximizing F1 on clean validation data, and freeze it across
all 28 conditions for that model (Table~\ref{tab:training}). Thresholds
are never re-tuned on corrupted data. A deployed system commits to one
threshold before a mission, so per-condition tuning would measure a best
case it does not have, and could credit a model with robustness that is
really a difference in how aggressively its threshold was re-optimized.
Because a frozen threshold could also exaggerate a failure, every
headline result is paired with threshold-independent mAP$_{50}$.

\paragraph{Shared evaluator.} All three architectures are scored by one
evaluator: greedy, score-ordered matching at IoU $\geq 0.5$, one
prediction per ground-truth box, class-agnostic. Predictions are
converted to a common format first, so no architecture is scored by its
own framework's metric. Numbers produced by three vendors' own evaluation
code are not comparable, and a buyer has no way to tell.

\paragraph{Tiling and full-image evaluation.} Both datasets pass through
one tiling pipeline (1024-pixel tiles, 256-pixel overlap). A HERIDAL frame
splits into 20 tiles; a SARD frame, smaller than one tile, passes through
whole. Boxes are mapped back to full-image coordinates and merged with
class-agnostic non-maximum suppression, so evaluation is always against
full-resolution ground truth. Corruptions are applied to the full frame
before tiling, since haze gradients and flood masks are spatially
coherent across a scene.

\paragraph{Localization stability.} For instances detected in both the
clean and the corrupted condition, we measure the IoU between the two
predicted boxes, a scale-normalized center shift, and the drop in
box-to-ground-truth IoU. Restricting to commonly detected instances keeps
the measure unconfounded by recall: it asks whether a surviving detection
stayed on the person, which is a different question from whether the
person was found at all, in the tradition of detector error
decompositions~\citep{hoiem2012diagnosing,bolya2020tide}.

\paragraph{Statistics.} Degradation is reported relative to each model's
own clean score, $\mathrm{rPD} = (\text{clean} -
\text{corrupted})/\text{clean}$, since the datasets do not share a
difficulty floor. Intervals are bootstrap 2.5th--97.5th percentiles over
1{,}000 resamples of test images~\citep{efron1993bootstrap}, so they
reflect test-set variance; training variance is addressed separately with
additional seeds (Section~\ref{sec:lowlight}).

\paragraph{Provenance and release.} Every result row logs the git commit,
configuration hash, global seed and timestamp, and raw predictions are
archived per condition, so any reported number traces to the code and
parameters that produced it. The corruption engine, evaluation pipeline,
configurations, a datasheet~\citep{gebru2021datasheets}, and the
per-condition results for every number in this paper are released under
the MIT license at \coderepo. HERIDAL and SARD imagery is not
redistributed and remains under its distributors' terms. An audit an
agency cannot rerun against a system it is considering is not much of an
audit.

\section{Experimental Setup}
\label{sec:setup}

\paragraph{Detectors.} We fine-tune three architecturally distinct
detectors from pretrained weights, one training run per model per
dataset: Faster R-CNN~\citep{ren2015fasterrcnn}, a two-stage CNN;
YOLOv11~\citep{ultralytics2024yolo11}, a single-stage CNN; and
RT-DETR~\citep{zhao2024rtdetr}, a transformer detector in the
DETR~\citep{carion2020detr} lineage. All three receive identical inputs
(1024-pixel tiles on HERIDAL, whole 640-pixel frames on SARD). YOLOv11
and RT-DETR resize them to a 1024-pixel network input, while Faster R-CNN
resizes internally to an 800-pixel shorter side, the torchvision default
(Table~\ref{tab:training}). Hyperparameters follow each
architecture's established fine-tuning practice rather than a shared
schedule, since a schedule tuned for one architecture is not fair to
another; the two-order-of-magnitude learning-rate gap between RT-DETR and
the CNNs follows that convention, and corrupted data enters neither
training nor threshold selection. Everything runs on a single consumer GPU
with 8~GB of memory, so the audit can be rerun without a cluster, which
matters if parties other than vendors are to rerun it.

\paragraph{Datasets.} The two datasets cover different capture regimes
(Table~\ref{tab:datasets}). HERIDAL~\citep{bozic2019heridal} contains
$4000\times3000$ aerial images of wilderness terrain captured from high
altitude, with persons occupying a tiny fraction of the frame. We use its
official 101-image test split untouched and carve a validation split
deterministically from the official training split.
SARD~\citep{sambolek2021sard} contains frames from drone video of actors
simulating injured or lost people across varied terrain, captured at
lower altitude. We use a Roboflow re-export resized to $640\times640$;
its documented preprocessing applies no augmentation, only
auto-orientation and the resize, yet it contains 5{,}755 images against
SARD's documented 1{,}981 source frames. Spot-checking colliding filenames
shows distinct frames, with different pixels and annotations, that share
a filename token, consistent with frame-counter reuse across the mirror's
upload history rather than duplicated or augmented content. Because
adjacent video frames are near-duplicates, splitting is group-aware: every
image sharing a filename token goes to exactly one split, the tooling
refuses degenerate groupings, and we verified that no group crosses
splits. Reproducing our SARD numbers requires the same re-export, since
the resize changes both tiling and the frame-relative corruption
parameters.

\begin{table}[t]
\centering\small
\caption{The two datasets as evaluated. Size bands follow COCO
(small $<32^2$, medium $32^2$--$96^2$, large $>96^2$ pixels of box area),
measured in the coordinates the corruptions are applied in.}
\label{tab:datasets}
\setlength{\tabcolsep}{5pt}
\begin{tabular}{lcc}
\toprule
 & HERIDAL & SARD \\
\midrule
Scene & Wilderness, high altitude & Staged rescue scenes, lower altitude \\
Evaluated resolution & $4000\times3000$ & $640\times640$ (re-export) \\
Tiles per frame (1024 / 256 overlap) & 20 & 1 (whole frame) \\
Images: train / val / test & 1{,}391 / 155 / 101 & 4{,}033 / 860 / 862 \\
Persons in test split & 337 & 1{,}097 \\
Test persons: small / medium / large & 5 / 327 / 5 & 155 / 789 / 153 \\
Split rule & Official test; val from train & Group-aware by filename token \\
\bottomrule
\end{tabular}
\end{table}

\section{Results}
\label{sec:results}

\subsection{Clean baselines}

Clean recall is 0.745 to 0.792 on HERIDAL and 0.879 to 0.891 on SARD
(Table~\ref{tab:s3}, first row; precision and mAP in
Table~\ref{tab:clean}). On HERIDAL our numbers sit modestly below the
commonly cited YOLOv5L reference point of roughly 0.90 precision, 0.893
recall and 0.834 mAP$_{50}$~\citep{bachir2024yolov5sar}, and are comparable
in precision and mAP; agreement at that level across different
architectures and input pipelines is all we claim, since clean accuracy is
not what this audit measures. These are the numbers a system is marketed
on, and every one of them is above 0.74.

\subsection{Robustness across conditions}

Robustness is highly uneven (Figure~\ref{fig:headline},
Table~\ref{tab:s3}). For YOLOv11 on SARD at severity 3, recall ranges from
0.855 under \cfd{rain\_streaks}, a 3.6-point drop from the 0.891 clean
baseline, to exactly 0.000 under \cfd{low\_light}. That range, produced by
corruptions each calibrated to its own declared ladder, is what makes the
benchmark diagnostic rather than merely hard. A single averaged robustness
score, the form a vendor is most likely to report, would compress the
whole finding into one number.

Table~\ref{tab:family} aggregates by disaster family. Flood is the mildest
family for every model on both datasets, the one ordering that holds
across the whole study. Storm is the worst family on HERIDAL and
earthquake dust the worst on SARD. Within families, the corruptions fail
in different ways. \cfd{water\_glare} costs recall moderately but is the
one corruption that also adds false alarms in bulk: on SARD at severity 3,
Faster R-CNN's precision falls from 0.967 to 0.513 as false positives
rise from 33 to 655, a cost a recall-only report would miss and one that
spends operator attention. The two Koschmieder hazes also separate
sharply: \cfd{dust\_haze} costs more recall than \cfd{smoke\_haze} in
all 18 model, dataset and severity comparisons, although their RMS
contrast is nearly identical and dust's optical depth is only 0.04 to 0.07
higher. Which part of the dust model carries the effect, airlight
chromaticity, transmission granularity, or grain, is not isolated here.

\begin{table}[t]
\centering\small
\caption{Recall at severity 3 for every corruption, detector and dataset, at
each model's frozen operating point. The clean row is the baseline each
column degrades from. The two corruptions whose parameters are lengths in
pixels (bottom block) reverse between datasets
(Section~\ref{sec:crossdataset}). Intervals and mAP$_{50}$ for every cell
are in Appendix~\ref{app:full}.}
\label{tab:s3}
\setlength{\tabcolsep}{4pt}
\begin{tabular}{llcccccc}
\toprule
 & & \multicolumn{3}{c}{HERIDAL} & \multicolumn{3}{c}{SARD} \\
\cmidrule(lr){3-5}\cmidrule(lr){6-8}
Corruption & Family & F.\,R-CNN & RT-DETR & YOLOv11 & F.\,R-CNN & RT-DETR & YOLOv11 \\
\midrule
Clean &  & 0.745 & 0.792 & 0.769 & 0.879 & 0.888 & 0.891 \\
\midrule
\cfd{water\_glare} & Flood & 0.614 & 0.682 & 0.653 & 0.629 & 0.646 & 0.666 \\
\cfd{turbidity\_cast} & Flood & 0.024 & 0.042 & 0.163 & 0.180 & 0.074 & 0.191 \\
\cfd{inundation} & Flood & 0.326 & 0.472 & 0.398 & 0.489 & 0.459 & 0.495 \\
\cfd{smoke\_haze} & Wildfire & 0.148 & 0.320 & 0.407 & 0.303 & 0.319 & 0.391 \\
\cfd{fire\_warm\_tint} & Wildfire & 0.401 & 0.086 & 0.145 & 0.407 & 0.118 & 0.099 \\
\cfd{dust\_haze} & Earthquake & 0.098 & 0.071 & 0.104 & 0.160 & 0.050 & 0.058 \\
\cfd{low\_light} & Storm & 0.000 & 0.000 & 0.000 & 0.000 & 0.000 & 0.000 \\
\midrule
\cfd{rain\_streaks} & Storm & 0.003 & 0.003 & 0.056 & 0.778 & 0.840 & 0.855 \\
\cfd{motion\_blur} & Storm & 0.000 & 0.000 & 0.003 & 0.398 & 0.329 & 0.478 \\
\bottomrule
\end{tabular}
\end{table}

\begin{table}[t]
\centering\small
\caption{Mean relative recall drop, $(\text{clean}-\text{corrupted})/\text{clean}$,
by disaster family, averaged over the family's corruptions and all three
severities. Flood is the mildest family for every model on both datasets.}
\label{tab:family}
\begin{tabular}{lcccccc}
\toprule
 & \multicolumn{3}{c}{HERIDAL} & \multicolumn{3}{c}{SARD} \\
\cmidrule(lr){2-4}\cmidrule(lr){5-7}
Family & F.\,R-CNN & RT-DETR & YOLOv11 & F.\,R-CNN & RT-DETR & YOLOv11 \\
\midrule
Flood & 0.301 & 0.231 & 0.246 & 0.264 & 0.293 & 0.252 \\
Wildfire & 0.335 & 0.459 & 0.405 & 0.327 & 0.451 & 0.445 \\
Earthquake & 0.534 & 0.568 & 0.571 & 0.481 & 0.630 & 0.624 \\
Storm & 0.784 & 0.874 & 0.829 & 0.401 & 0.440 & 0.413 \\
\bottomrule
\end{tabular}
\end{table}

\subsection{Spatial corruptions do not transfer across capture regimes}
\label{sec:crossdataset}

Corruptions that act on intensity or colour cost broadly similar amounts
on the two datasets. Averaged by family, the relative drops for flood,
wildfire and earthquake agree across datasets to within 0.07 for every
model (Table~\ref{tab:family}); at severity 3, \cfd{water\_glare} leaves
recall at 0.614--0.682 on HERIDAL and 0.629--0.666 on SARD, and
\cfd{low\_light} reaches zero on both. Storm, the family that contains
both corruptions defined by a length in pixels, does not agree (0.78--0.87
against 0.40--0.44), and those two corruptions are why.
\cfd{rain\_streaks} and \cfd{motion\_blur} are mild to moderate on SARD
(0.778--0.855 and 0.329--0.478 at severity 3) and near-total on HERIDAL
(0.003--0.056 and 0.000--0.003). On HERIDAL, motion blur already removes
almost every detection at severity 1 (recall 0.003 to 0.107).

This reversal should not be read as an altitude effect. Spatial
parameters scale with the frame, so a severity-3 blur kernel is 100
pixels on HERIDAL and 16 on SARD, while most persons in both datasets fall
in the same 32--96-pixel size band (Table~\ref{tab:datasets}). Relative to
the people being searched for, the HERIDAL kernel is several times longer,
which predicts the observed pattern without any appeal to altitude. We
have not run the target-relative control that would separate the two, so
we report the reversal as a measured property of this benchmark's
geometry, not of the capture regimes. The practical point holds either
way: a robustness figure for a spatial degradation is not portable between
imaging configurations, and should be reported together with the image
size and target size it was measured at.

\subsection{Detection collapse under low light}
\label{sec:lowlight}

\cfd{low\_light} is the audit's headline result
(Table~\ref{tab:lowlight}). At severity 3, recall is exactly 0.000 for all
three architectures on both HERIDAL (101 images, 337 persons) and SARD
(862 images, 1{,}097 persons), and every one of 1{,}000 bootstrap resamples
agrees, so each interval is degenerate at $[0.000, 0.000]$. A two-stage
CNN, a single-stage CNN, and a transformer are about as architecturally
different as current detectors get, and all three reach the same floor.
That rules out an architecture-specific explanation and means an agency
cannot buy its way out of the problem by choosing a different vendor's
model.

\begin{table}[t]
\centering\small
\caption{Recall under \cfd{low\_light} at each model's frozen operating point
(bootstrap 95\% intervals, 1{,}000 image resamples), and threshold-independent
mAP$_{50}$. Severities 1, 2 and 3 are exposure losses of 1.60, 2.94 and 4.47
stops. Once recall is exactly zero on the full test set every resample
returns zero, so the interval is degenerate by construction.}
\label{tab:lowlight}
\footnotesize\setlength{\tabcolsep}{2.6pt}
\begin{tabular}{llccccccc}
\toprule
 & & \multicolumn{3}{c}{Recall [95\% CI]} & \multicolumn{4}{c}{mAP$_{50}$} \\
\cmidrule(lr){3-5}\cmidrule(lr){6-9}
Model & Thr. & s1 & s2 & s3 & clean & s1 & s2 & s3 \\
\midrule
\multicolumn{9}{l}{\emph{HERIDAL ($n=101$ images, 337 persons)}} \\
Faster R-CNN & 0.99 & 0.451 [0.365, 0.540] & 0.110 [0.064, 0.163] & \textbf{0.000} [0.000, 0.000] & 0.785 & 0.537 & 0.176 & 0.010 \\
RT-DETR & 0.65 & 0.000 [0.000, 0.000] & 0.000 [0.000, 0.000] & \textbf{0.000} [0.000, 0.000] & 0.835 & 0.001 & 0.000 & 0.000 \\
YOLOv11 & 0.45 & 0.047 [0.027, 0.073] & 0.000 [0.000, 0.000] & \textbf{0.000} [0.000, 0.000] & 0.803 & 0.064 & 0.000 & 0.000 \\
\midrule
\multicolumn{9}{l}{\emph{SARD ($n=862$ images, 1{,}097 persons)}} \\
Faster R-CNN & 0.96 & 0.419 [0.389, 0.452] & 0.077 [0.060, 0.094] & \textbf{0.000} [0.000, 0.000] & 0.896 & 0.546 & 0.142 & 0.000 \\
RT-DETR & 0.60 & 0.260 [0.233, 0.288] & 0.013 [0.007, 0.019] & \textbf{0.000} [0.000, 0.000] & 0.929 & 0.421 & 0.049 & 0.001 \\
YOLOv11 & 0.30 & 0.184 [0.157, 0.212] & 0.007 [0.003, 0.012] & \textbf{0.000} [0.000, 0.000] & 0.928 & 0.247 & 0.018 & 0.000 \\
\bottomrule
\end{tabular}
\end{table}

\paragraph{Not an operating-point artifact.} A frozen-threshold zero could
reflect a badly placed threshold rather than missing detections.
mAP$_{50}$ settles this, since it integrates over every threshold. On
HERIDAL at severity 1, RT-DETR's mAP$_{50}$ falls from 0.835 clean to
0.001, YOLOv11's from 0.803 to 0.064, and Faster R-CNN's from 0.785 to
0.537, in step with frozen-threshold recall, and by severity 3 no model on
either dataset exceeds 0.010. If the threshold were the problem, mAP would
stay high while recall fell. It does not, so the detections are absent,
not hidden below the threshold.

Faster R-CNN is the most robust model at low light wherever any model
retains recall, despite having the strictest threshold (0.99 on HERIDAL),
so threshold strictness does not explain the ordering. The order of
YOLOv11 and RT-DETR reverses between datasets: RT-DETR finds nobody on
HERIDAL even at severity 1 but beats YOLOv11 at severities 1--2 on SARD.
We report that as dataset-dependent rather than as a property of either
architecture.

\paragraph{Not a property of one training run.} We retrained the SARD
YOLOv11 at two further seeds, each with its own threshold selected by the
same rule (Table~\ref{tab:seeds}). Clean recall varies little across the
three runs (0.874 to 0.891) and severity-1 recall varies more (0.184 to
0.234), which is the honest scale of run-to-run variation on this ladder.
All three return exactly 0.000 at severity 3.

\begin{table}[t]
\centering\small
\caption{Three independently trained YOLOv11 checkpoints on SARD, each with
its own threshold selected by the same rule. Recall under \cfd{low\_light}
with bootstrap 95\% intervals.}
\label{tab:seeds}
\setlength{\tabcolsep}{4pt}
\begin{tabular}{cccccc}
\toprule
Seed & Thr. & Clean & s1 & s2 & s3 \\
\midrule
17 & 0.30 & 0.891 & 0.184 [0.157, 0.212] & 0.007 [0.003, 0.012] & \textbf{0.000} [0.000, 0.000] \\
42 & 0.35 & 0.874 & 0.200 [0.170, 0.229] & 0.008 [0.003, 0.014] & \textbf{0.000} [0.000, 0.000] \\
123 & 0.30 & 0.885 & 0.234 [0.207, 0.265] & 0.021 [0.012, 0.029] & \textbf{0.000} [0.000, 0.000] \\
\bottomrule
\end{tabular}
\end{table}

\paragraph{Is the signal still in the image?} A zero could mean the
detector fails, or that the person is no longer recoverable from the
pixels. Three observations bear on this, and none settles it completely.
First, severity 3 is a 4.47-stop exposure loss, a dim dusk exposure rather
than darkness. Second, at severity 1 on HERIDAL, Faster R-CNN still finds
0.451 of persons on the same images where RT-DETR finds none, so at that
level the information is present and the difference is the detector.
Third, targeted augmentation recovers part of the held-out severity
(below), so some usable signal survives even at severity 3. The recovery
is partial, so a real share of the loss may be a sensor limit rather than
a detector limit. Where visible-band signal does run out, thermal infrared
is the natural complementary modality (Section~\ref{sec:deployment}).

\paragraph{A check against real night imagery.} Because the corruption is
synthetic, we evaluated the SARD-trained YOLOv11, at its frozen threshold
and without retraining, on 300 real night images from the VisDrone-DET
test-dev split~\citep{zhu2022visdrone}: the 300 lowest-luminance images,
each confirmed by eye as dusk or night, scored against VisDrone's own
pedestrian and people annotations. The model found 2 of 759 annotated
persons (recall 0.003), close to the floor our synthetic \cfd{low\_light}
reaches at severity 3. This is consistent with the synthetic result but
does not isolate darkness: VisDrone night scenes are urban and
artificially lit, and its pedestrians differ from SARD's in scene and
scale, so part of this failure is ordinary domain shift. A matched
daytime VisDrone control would separate the two, and we flag it as the
first check to add.

\paragraph{Targeted augmentation moves the floor, at a price.} We
fine-tuned each SARD detector from its clean checkpoint on a training set
in which every image also appears once with \cfd{low\_light} at severity 1
or 2, drawn uniformly; severity 3 never appears in training, and
validation stays clean. Each model then had its threshold re-selected on
clean validation data by the same rule and was re-evaluated on all 28
conditions (Table~\ref{tab:mitigation}). Held-out severity-3 recall rises
from zero to 0.121 [0.101, 0.140] for Faster R-CNN, 0.059 [0.046, 0.073]
for RT-DETR, and 0.188 [0.168, 0.211] for YOLOv11, and severity-3
mAP$_{50}$ rises from below 0.002 to 0.152, 0.134 and 0.368. The floor
moves for all three at a severity never seen in training, so the collapse
is at least partly a coverage gap in the training data rather than a hard
architectural ceiling. It is not an operational fix: finding between one
in five and one in seventeen people is still a failed search.

The cost elsewhere depends on the architecture, and in opposite
directions. Across the 24 corrupted conditions outside \cfd{low\_light},
Faster R-CNN \emph{improves} on average ($+$0.037, none worse by more than
0.03), YOLOv11 loses a little ($-$0.019, 8 of 24 worse by more than 0.03),
and RT-DETR pays a real tax ($-$0.043, 17 of 24 worse by more than 0.03,
and $-$0.030 on clean imagery). RT-DETR already had the largest relative drop of the three in every
disaster family on SARD, and it is also the model for which this
augmentation costs the most. An averaged robustness score would hide this
interaction completely. Finally, the augmented YOLOv11 found 4 of the 759
persons in the real night images, against 2 before: the augmentation that
moves the synthetic floor does not move the real one.

\begin{table}[t]
\centering\small
\caption{Low-light augmentation on SARD. Each clean checkpoint (\emph{base})
is fine-tuned on training images augmented with \cfd{low\_light} at
severities 1--2 (\emph{aug}); severity 3 never appears in training.
Thresholds are re-selected on clean validation data by the same rule.
Recall at the frozen operating point, with the bootstrap 95\% interval at
the held-out severity. Right: change in recall over the 24 corrupted
conditions outside \cfd{low\_light}, computed before rounding.}
\label{tab:mitigation}
\footnotesize\setlength{\tabcolsep}{3pt}
\begin{tabular}{llcccccc@{\hspace{8pt}}ccc}
\toprule
 & & & & \multicolumn{4}{c}{\cfd{low\_light}} & \multicolumn{3}{c}{Other 24 conditions} \\
\cmidrule(lr){5-8}\cmidrule(lr){9-11}
Model & & Thr. & Clean & s1 & s2 & s3 (held out) & mAP$_{50}$ s3 & Mean $\Delta$ & Worst $\Delta$ & $\Delta<-0.03$ \\
\midrule
Faster R-CNN & base & 0.96 & 0.879 & 0.419 & 0.077 & 0.000 [0.000, 0.000] & 0.000 &  &  &  \\
 & aug & 0.65 & 0.900 & 0.843 & 0.641 & \textbf{0.121} [0.101, 0.140] & 0.152 & $+$0.037 & $-$0.025 & 0/24 \\
\midrule
RT-DETR & base & 0.60 & 0.888 & 0.260 & 0.013 & 0.000 [0.000, 0.000] & 0.001 &  &  &  \\
 & aug & 0.55 & 0.858 & 0.772 & 0.529 & \textbf{0.059} [0.046, 0.073] & 0.134 & $-$0.043 & $-$0.116 & 17/24 \\
\midrule
YOLOv11 & base & 0.30 & 0.891 & 0.184 & 0.007 & 0.000 [0.000, 0.000] & 0.000 &  &  &  \\
 & aug & 0.30 & 0.883 & 0.807 & 0.647 & \textbf{0.188} [0.168, 0.211] & 0.368 & $-$0.019 & $-$0.066 & 8/24 \\
\bottomrule
\end{tabular}
\end{table}

\subsection{Localization stability: the failure is silent}
\label{sec:locstab}

\begin{figure}[t]
\centering
\begin{minipage}[b]{0.31\linewidth}
\centering
\includegraphics[width=\linewidth]{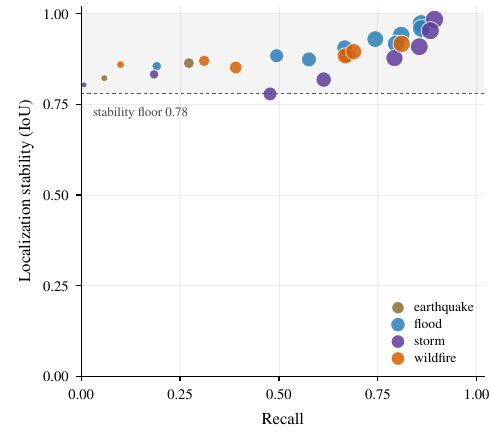}
\end{minipage}\hfill
\begin{minipage}[b]{0.66\linewidth}
\centering
\includegraphics[width=\linewidth]{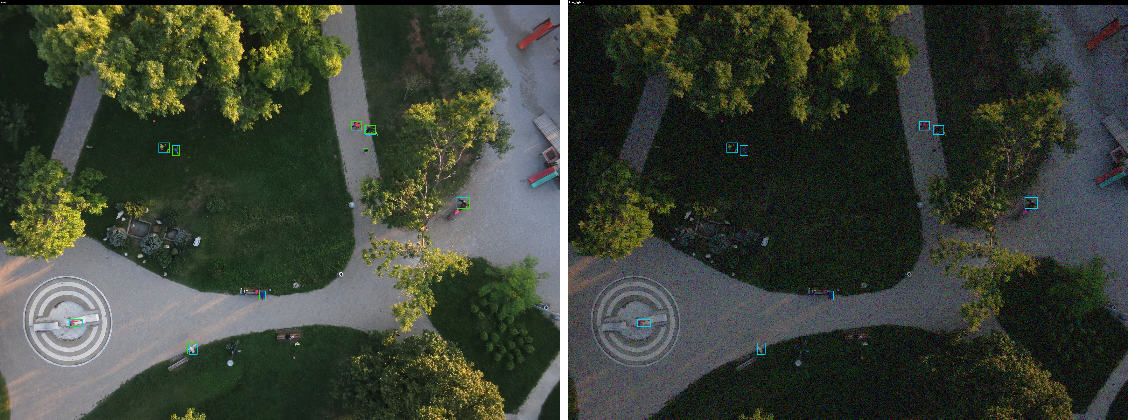}
\end{minipage}
\caption{\textbf{Left:} recall against localization stability, YOLOv11 on
SARD, one point per corruption and severity, point area proportional to
the number of persons detected in both conditions. Points sweep the whole
recall range while staying in a narrow band near the top of the stability
axis. \textbf{Right:} a clean HERIDAL frame with ground truth (cyan) and
retained detections (green), beside the same frame under \cfd{low\_light}
at severity 1, the mildest level in the benchmark. Every detection is
lost, and nothing in the output says so.}
\label{fig:decoupling}
\end{figure}

Figure~\ref{fig:decoupling} plots recall against localization stability
for YOLOv11 on SARD, one point per corruption and severity. Both axes span
the full unit range on purpose. Stability declines from 0.98 to 0.78 while
recall falls from 0.89 to 0.00, so both decline and we do not claim they
are independent. The claim the data supports is about magnitude: recall
sweeps almost the entire unit range while stability stays in a narrow
band near the top, never below 0.78 even where nearly every detection is
lost. Meanwhile $n_{\text{common}}$, the number of persons detected in
both conditions, tracks recall exactly (\cfd{low\_light}: 202, then 8,
then 0). The dominant failure is a detection disappearing, not a box
drifting off the person. Faster R-CNN and RT-DETR on SARD show the same
band (0.81--0.98 and 0.77--0.99). The exception is spatial blur on
HERIDAL, where the few detections that survive do drift (YOLOv11 under
\cfd{motion\_blur} at severity 1: stability 0.62 over 34 surviving
detections), which is what a smeared image should do.
Figure~\ref{fig:decoupling} shows the low-light case directly, and
Figure~\ref{fig:dusthaze} in the appendix shows a partial-degradation
case.

\section{Implications for Deployment and Reporting}
\label{sec:deployment}

The measurements above are computer-vision results; their consequences
are operational, and three of them generalize past this task.

\paragraph{The failure mode defeats human oversight.} Keeping a human in
the loop works when errors are visible. Here they mostly are not:
detections are lost rather than misplaced, so an operator reviewing a
night-flight feed with no detections has no signal distinguishing ``no one
is down there'' from ``this system cannot see anyone right now.''
Oversight that depends on noticing a mistake does not work against a
failure that produces no visible mistake. That is a general property of
recall-side failures, not a quirk of this setup.

\paragraph{Clear-condition benchmarks are the wrong evidence standard.}
Every model here has clean recall above 0.74, and every model reaches
exactly zero recall at a 4.5-stop exposure loss. Both numbers are true and
usually only the first is published, so an agency comparing systems on
published accuracy is deciding on evidence that does not address its
question. Unusually among applications with this kind of gap, the
deployment conditions are physically characterizable, so the gap can be
closed by measurement rather than only flagged.

\paragraph{A reporting practice.} Model cards already ask for results
disaggregated by relevant conditions~\citep{mitchell2019modelcards}. For
aerial search and rescue we suggest the conditions be environmental, and
that a report include, at minimum:
\begin{enumerate}[leftmargin=1.6em,itemsep=1pt,topsep=2pt]
\item recall per named operating condition (night, smoke, dust, flood
  water, rain) at the single threshold the system will actually fly with;
\item a threshold-independent metric beside it, so a calibration problem
  can be told apart from missing detections;
\item for spatial degradations such as blur and rain, the image size and
  typical target size in pixels the number was measured at, since it does
  not transfer between them (Section~\ref{sec:crossdataset});
\item precision as well as recall under each condition, since some
  conditions (glare) fail partly by adding false alarms, not only by losing
  people;
\item for any robustness fix, its effect on the full condition grid and on
  real imagery, not only on the condition it targets
  (Section~\ref{sec:lowlight}).
\end{enumerate}
The released pipeline reports the first four for any detector wrapped
behind its evaluator, and the mitigation tooling covers the fifth.

\paragraph{What follows operationally.} We do not claim these systems
should not be used; in daylight they help, and the alternative in a
wide-area search is often no coverage at all. What follows is narrower. A
system with near-zero recall in a condition should not be the primary
search method in that condition, operators should be told which
conditions those are, and the condition should be detected at flight time.
Ambient light is directly measurable on the aircraft, so for the
headline failure this guardrail is cheap. Targeted augmentation moves the
synthetic floor but not the real one, and its cost elsewhere depends on
the architecture, so it has to be validated on the full grid and on real
imagery before it is trusted. Where the visible-band signal itself runs
out, the answer is a different sensor: thermal infrared does not depend
on ambient light and is the natural complement for night search.

\section{Limitations and Open Questions}
\label{sec:limitations}

Our corruptions are synthetic, applied to real imagery in the ImageNet-C
and Foggy Cityscapes
tradition~\citep{hendrycks2019imagenetc,sakaridis2018foggy}. That is what
makes controlled, per-condition attribution possible, since disaster
imagery almost never comes with a matched clean view of the same scene,
and it is also the ceiling: we measure response to modeled optics, not to
live disaster footage. Severity levels are monotone in stated physical
quantities but not fitted to field measurements. Corruptions are applied
one at a time, while a real night flood combines several. We evaluate two
datasets and three detectors, and the seed and augmentation studies cover
SARD only. \cfd{rain\_streaks}' severity is defined by generation
parameters rather than a measured statistic.

The results leave four questions open, none of which needs new training:
\begin{enumerate}[leftmargin=1.6em,itemsep=1pt,topsep=2pt]
\item \textbf{Geometry.} Does the rain and blur reversal survive when
  spatial parameters are set relative to target size instead of frame
  size (Section~\ref{sec:crossdataset})?
\item \textbf{Low-light mechanism.} Which component of the low-light
  model, photon gain, shot and read noise, or the blue shift, drives
  recall to zero? Ablating them at matched mean luminance would tell
  whether the failure is about brightness or about noise.
\item \textbf{Smoke versus dust.} Swapping the two hazes' airlight colours
  while holding their transmission fields fixed would show whether
  chromaticity explains why dust costs more.
\item \textbf{Real night imagery.} A matched daytime control on the same
  VisDrone scenes would separate darkness from domain shift in the
  real-night check (Section~\ref{sec:lowlight}).
\end{enumerate}

\section{Conclusion}
\label{sec:conclusion}

Aerial search-and-rescue person detection is benchmarked on calm, clear
imagery and deployed into disasters. AegisBench measures that gap with
physically grounded, calibrated corruptions and a protocol that scores
detectors the way they are fielded. Robustness is highly uneven, and low
light collapses recall to exactly zero for three architecturally distinct
detectors on both datasets, a result that survives a threshold-free check
and additional training seeds. Because surviving detections largely stay
on target, the failure is silent. Targeted augmentation moves the floor
only partly, at a cost that depends on the architecture. Systems intended
for disaster response should be evaluated under disaster conditions,
reported by condition, and restricted at flight time where they do not
work. The benchmark, the pipeline, and every number in this paper are
released so that anyone procuring or building these systems can run the
same audit.

Three groups can use this directly. Agencies and vendors can adopt the
five-item reporting practice of Section~\ref{sec:deployment} as a
procurement requirement, so that a system is judged on the conditions it
will fly in and not on a clear-sky score. Researchers get a public
168-condition baseline, a corruption engine that regenerates the test sets
bit-identically, and an evaluator any detector can be wrapped behind, all
runnable on a single 8~GB GPU. And the central finding, a failure that
removes detections without producing a visible error, is not specific to
this task: any safety-of-life detector whose misses are silent needs a
condition-stratified, frozen-threshold evaluation before its performance is
trusted. The cheapest gain is also the most immediate. Ambient light is
measurable on the aircraft, so a flight-time check that tells operators
when the system cannot see costs almost nothing and turns a silent failure
into a visible one. We use only public research datasets under their own
terms, collect no human-subjects data, and introduce no new detector. None
of this argues against aerial detection, which in daylight extends coverage
ground teams cannot provide; it argues that the conditions where the
technology matters most must be measured before it is relied on there.

\section*{Use of Generative AI}

Generative AI tools helped with drafting the text of this manuscript. All
experiments, evaluations and results were run and verified by the authors,
who take full responsibility for the content.

\bibliographystyle{unsrtnat}
\bibliography{aegisbench}

\appendix
\section{Clean baselines and training details}
\label{app:tables}

\begin{table}[h]
\centering\small
\caption{Clean-condition performance at each model's frozen operating
point. All degradation results are relative to these baselines, since the
two datasets do not share a difficulty floor.}
\label{tab:clean}
\begin{tabular}{lcccc}
\toprule
Model & Recall & Precision & mAP$_{50}$ & mAP$_{50:95}$ \\
\midrule
\multicolumn{5}{l}{\emph{HERIDAL}} \\
Faster R-CNN & 0.745 & 0.903 & 0.785 & 0.518 \\
RT-DETR      & 0.792 & 0.881 & 0.835 & 0.556 \\
YOLOv11      & 0.769 & 0.855 & 0.803 & 0.510 \\
\midrule
\multicolumn{5}{l}{\emph{SARD}} \\
Faster R-CNN & 0.879 & 0.967 & 0.896 & 0.615 \\
RT-DETR      & 0.888 & 0.972 & 0.929 & 0.633 \\
YOLOv11      & 0.891 & 0.955 & 0.928 & 0.601 \\
\bottomrule
\end{tabular}
\end{table}

\begin{table}[t]
\centering\small
\caption{Training configuration and frozen operating points. Hyperparameters
follow established per-architecture fine-tuning practice rather than a
shared schedule. Thresholds are selected once per model per dataset by
maximizing F1 on clean validation data and held fixed across all 28
conditions. Corrupted data enters neither training nor threshold selection.
All three models receive identical tiles; Faster R-CNN resizes them
internally to an 800-pixel shorter side (torchvision default).
Bottom block: the low-light fine-tunes of Section~\ref{sec:lowlight}, each
started from the clean SARD checkpoint.}
\label{tab:training}
\setlength{\tabcolsep}{4pt}
\begin{tabular}{lccc}
\toprule
 & YOLOv11 & RT-DETR & Faster R-CNN \\
\midrule
Base weights   & \cfd{yolo11m} (COCO) & \cfd{rtdetr-l} (COCO) & R50-FPN v2 (COCO) \\
Epochs         & 80     & 80      & 26     \\
Batch size     & 4      & 2       & 2      \\
Input size     & 1024   & 1024    & 800 (internal) \\
Initial LR     & 0.01   & 0.0001  & 0.005  \\
Early-stop patience & 20 & 20     & ---    \\
Seed           & 17     & 17      & 17     \\
\midrule
Threshold, HERIDAL & 0.45 & 0.65 & 0.99 \\
Threshold, SARD    & 0.30 & 0.60 & 0.96 \\
\midrule
\multicolumn{4}{l}{\emph{Low-light fine-tune (SARD)}} \\
Epochs / patience  & 20 / 8 & 20 / 8 & 8 / --- \\
Initial LR         & 0.001  & 0.00001 & 0.0005 \\
Mixed precision    & on     & off     & on \\
Threshold          & 0.30   & 0.55    & 0.65 \\
\bottomrule
\end{tabular}
\end{table}

\paragraph{Low-light fine-tuning.} The augmented SARD training set keeps
every clean training image and adds one copy corrupted with
\cfd{low\_light}, its severity drawn uniformly from $\{1, 2\}$ with the
same deterministic seeding as the benchmark, so augmented images make up
half of the training set. Validation images stay clean, so threshold
selection is unaffected by the augmentation. Each model starts from its
clean SARD checkpoint (Table~\ref{tab:training}, bottom block).

\section{Corruption models}
\label{app:models}

Table~\ref{tab:params} lists every parameter of every corruption. A few
modelling notes follow.

\paragraph{Haze family.} All three hazes composite the clear image toward
an airlight colour through a spatially varying transmission field built
from multi-octave value noise around the severity's mean transmission.
\cfd{smoke\_haze} uses a neutral-gray airlight and a smooth, low-frequency
field. \cfd{dust\_haze} uses a brown mineral airlight, a patchier
higher-frequency field, and coarse additive grain modelling near-lens
particulates. \cfd{fire\_warm\_tint} first shifts white balance toward
low-colour-temperature firelight (red gain up, blue gain down) and then
applies a lighter haze with a warm airlight.

\paragraph{Flood family.} \cfd{turbidity\_cast} blends toward a mud
colour, compresses contrast toward the scene mean, and reduces saturation.
\cfd{inundation} selects a flooded region with a feathered noise mask and,
inside it, blends toward murky water, adds a small sinusoidal refraction
ripple, and sprinkles specular sparkle. \cfd{water\_glare} thresholds a
noise field into glint cores and screen-blends them with a Gaussian bloom,
saturating pixels the way glint saturates a sensor.

\paragraph{Storm family.} \cfd{low\_light} decodes to approximately linear
intensity (gamma 2.2), scales by the severity's photon gain, applies a
twilight white-balance shift (red gain 0.92, blue gain 1.06), adds
signal-dependent shot noise and constant read noise, and re-encodes.
\cfd{rain\_streaks} first applies a rain veil (slight blur, desaturation,
darkening) and then adds streaks along one wind direction per image with
small per-streak jitter. \cfd{motion\_blur} convolves with a linear kernel
in one random direction per image.

\begin{table}[p]
\centering\scriptsize
\caption{Every corruption parameter, as set in the configuration file
\cfd{configs/corruptions.yaml} of the code release. Parameters ending in \cfd{per\_1000px} are
scaled at runtime by the longer image side divided by 1000. Global
corruption seed: 20260707.}
\label{tab:params}
\setlength{\tabcolsep}{3pt}
\begin{tabular}{llccc}
\toprule
Corruption & Parameter & s1 & s2 & s3 \\
\midrule
\cfd{water\_glare} & \cfd{coverage} & 0.03 & 0.08 & 0.18 \\
 & \cfd{intensity} & 0.7 & 0.85 & 1 \\
 & \cfd{bloom\_sigma\_per\_1000px} & 8 & 12 & 16 \\
 & \multicolumn{4}{>{\raggedright\arraybackslash}p{0.72\linewidth}}{\emph{Fixed:} \cfd{glare\_color\_rgb} (1, 0.98, 0.92), \cfd{noise\_octaves} 4, \cfd{noise\_base\_res} 6} \\
\midrule
\cfd{turbidity\_cast} & \cfd{mud\_alpha} & 0.15 & 0.3 & 0.48 \\
 & \cfd{contrast\_factor} & 0.8 & 0.62 & 0.45 \\
 & \cfd{saturation\_factor} & 0.9 & 0.75 & 0.6 \\
 & \multicolumn{4}{>{\raggedright\arraybackslash}p{0.72\linewidth}}{\emph{Fixed:} \cfd{mud\_color\_rgb} (0.42, 0.33, 0.23)} \\
\midrule
\cfd{inundation} & \cfd{coverage} & 0.2 & 0.4 & 0.6 \\
 & \cfd{opacity} & 0.5 & 0.65 & 0.8 \\
 & \cfd{ripple\_amp\_per\_1000px} & 0.15 & 0.25 & 0.35 \\
 & \multicolumn{4}{>{\raggedright\arraybackslash}p{0.72\linewidth}}{\emph{Fixed:} \cfd{water\_color\_rgb} (0.35, 0.38, 0.3), \cfd{ripple\_wavelength\_per\_1000px} 60, \cfd{mask\_feather\_per\_1000px} 15, \cfd{sparkle\_intensity} 0.25, \cfd{noise\_octaves} 3, \cfd{noise\_base\_res} 4} \\
\midrule
\cfd{smoke\_haze} & \cfd{t\_mean} & 0.7 & 0.45 & 0.25 \\
 & \multicolumn{4}{>{\raggedright\arraybackslash}p{0.72\linewidth}}{\emph{Fixed:} \cfd{airlight\_rgb} (0.78, 0.78, 0.8), \cfd{t\_variation\_amp} 0.15, \cfd{noise\_octaves} 3, \cfd{noise\_base\_res} 3} \\
\midrule
\cfd{fire\_warm\_tint} & \cfd{r\_gain} & 1.08 & 1.18 & 1.3 \\
 & \cfd{b\_gain} & 0.92 & 0.82 & 0.7 \\
 & \cfd{t\_mean} & 0.85 & 0.72 & 0.58 \\
 & \multicolumn{4}{>{\raggedright\arraybackslash}p{0.72\linewidth}}{\emph{Fixed:} \cfd{airlight\_rgb} (0.91, 0.62, 0.35), \cfd{t\_variation\_amp} 0.25, \cfd{noise\_octaves} 4, \cfd{noise\_base\_res} 5} \\
\midrule
\cfd{rain\_streaks} & \cfd{streaks\_per\_megapixel} & 120 & 350 & 800 \\
 & \cfd{length\_per\_1000px} & 12 & 20 & 30 \\
 & \cfd{veil\_blur\_sigma\_per\_1000px} & 0.4 & 0.8 & 1.2 \\
 & \cfd{brightness\_factor} & 0.95 & 0.88 & 0.8 \\
 & \cfd{saturation\_factor} & 0.92 & 0.85 & 0.75 \\
 & \multicolumn{4}{>{\raggedright\arraybackslash}p{0.72\linewidth}}{\emph{Fixed:} \cfd{angle\_mean\_deg} 70, \cfd{angle\_jitter\_image\_deg} 8, \cfd{angle\_jitter\_streak\_deg} 3, \cfd{streak\_intensity\_range} (0.15, 0.45)} \\
\midrule
\cfd{motion\_blur} & \cfd{kernel\_len\_per\_1000px} & 9 & 17 & 25 \\
\midrule
\cfd{low\_light} & \cfd{linear\_gain} & 0.33 & 0.13 & 0.045 \\
 & \cfd{shot\_noise\_coef} & 0.012 & 0.02 & 0.03 \\
 & \cfd{read\_noise\_sigma} & 0.01 & 0.018 & 0.028 \\
 & \multicolumn{4}{>{\raggedright\arraybackslash}p{0.72\linewidth}}{\emph{Fixed:} \cfd{wb\_r\_gain} 0.92, \cfd{wb\_b\_gain} 1.06, \cfd{decode\_gamma} 2.2} \\
\midrule
\cfd{dust\_haze} & \cfd{t\_mean} & 0.65 & 0.42 & 0.24 \\
 & \cfd{grain\_amp} & 0.02 & 0.03 & 0.04 \\
 & \multicolumn{4}{>{\raggedright\arraybackslash}p{0.72\linewidth}}{\emph{Fixed:} \cfd{airlight\_rgb} (0.7, 0.56, 0.38), \cfd{t\_variation\_amp} 0.3, \cfd{noise\_octaves} 5, \cfd{noise\_base\_res} 8, \cfd{grain\_base\_res} 220} \\
\bottomrule
\end{tabular}
\end{table}

\section{Additional figures}
\label{app:figures}

\begin{figure}[h]
\centering
\includegraphics[width=\linewidth]{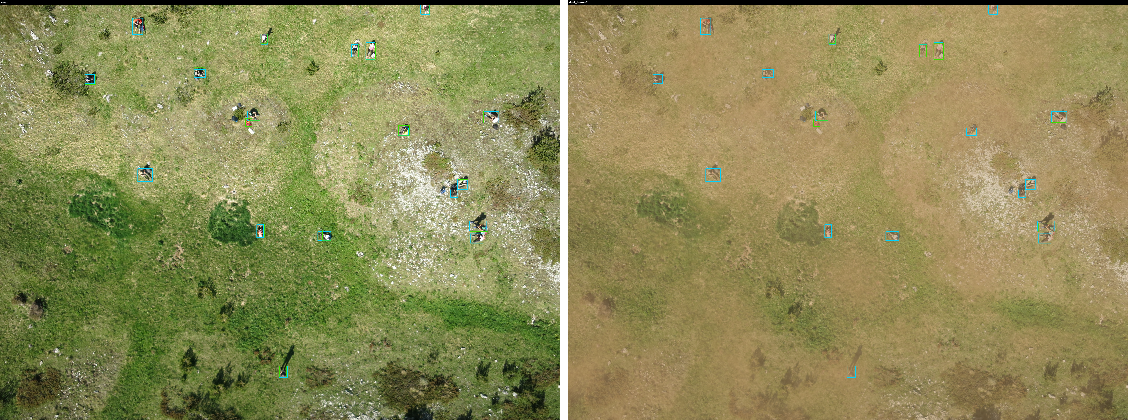}
\caption{Partial degradation under \cfd{dust\_haze}. A clean HERIDAL frame
(RT-DETR, test image \cfd{BLI\_0004}) beside the same frame under
\cfd{dust\_haze} at severity 2. RT-DETR finds 16 of 18 annotated people on
the clean frame and 7 of 18 under corruption. Ground truth in cyan,
detections retained at the frozen operating point in green.}
\label{fig:dusthaze}
\end{figure}

\begin{figure}[t]
\centering
\includegraphics[width=\linewidth]{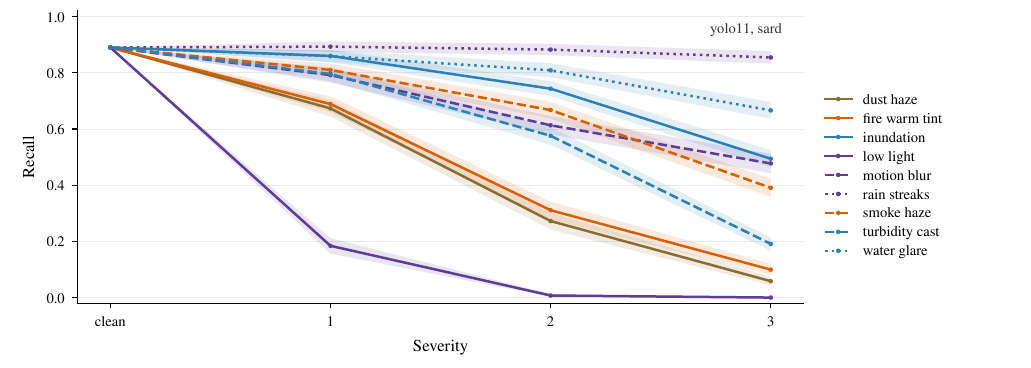}
\caption{Recall against corruption severity for YOLOv11 on SARD, with
bootstrap 95\% intervals shaded (1{,}000 image resamples per condition).
All curves start from the same clean baseline. Colour indicates disaster
family; line style distinguishes corruptions within a family.}
\label{fig:severity}
\end{figure}

Figure~\ref{fig:dusthaze} shows a partial-degradation case, which the
\cfd{low\_light} pair in Figure~\ref{fig:decoupling} does not: detections
are lost progressively rather than all at once, and the boxes that survive
stay on the person. Figure~\ref{fig:severity} shows every severity ladder
for YOLOv11 on SARD with its bootstrap band.

\section{Full results}
\label{app:full}

Tables~\ref{tab:full_heridal} and~\ref{tab:full_sard} give recall with its
bootstrap interval and mAP$_{50}$ for all 168 conditions. The same values,
with precision, F1, mAP$_{50:95}$ and true and false positive counts, are
in \cfd{results/workshop/benchmark\_168\_conditions.csv} in the code
release; size-stratified recall is in \cfd{results/sweep/master\_ci.csv}.

\begin{table}[p]
\centering\scriptsize
\caption{Full results, HERIDAL. Recall at the frozen operating point with bootstrap 95\% interval (1{,}000 image resamples), and threshold-independent mAP$_{50}$. Thresholds: Faster R-CNN 0.99, RT-DETR 0.65, YOLOv11 0.45.}
\label{tab:full_heridal}
\setlength{\tabcolsep}{3pt}
\begin{tabular}{lccccccc}
\toprule
 & & \multicolumn{2}{c}{Faster R-CNN} & \multicolumn{2}{c}{RT-DETR} & \multicolumn{2}{c}{YOLOv11} \\
\cmidrule(lr){3-4}\cmidrule(lr){5-6}\cmidrule(lr){7-8}
Corruption & Sev. & Recall [95\% CI] & mAP$_{50}$ & Recall [95\% CI] & mAP$_{50}$ & Recall [95\% CI] & mAP$_{50}$ \\
\midrule
Clean & -- & 0.745 [0.664, 0.829] & 0.785 & 0.792 [0.718, 0.873] & 0.835 & 0.769 [0.702, 0.838] & 0.803 \\
\midrule
\cfd{water\_glare} & 1 & 0.697 [0.621, 0.784] & 0.742 & 0.769 [0.698, 0.847] & 0.834 & 0.739 [0.667, 0.817] & 0.785 \\
 & 2 & 0.706 [0.634, 0.788] & 0.745 & 0.777 [0.707, 0.850] & 0.832 & 0.733 [0.669, 0.803] & 0.781 \\
 & 3 & 0.614 [0.523, 0.708] & 0.661 & 0.682 [0.607, 0.753] & 0.725 & 0.653 [0.582, 0.728] & 0.680 \\
\midrule
\cfd{turbidity\_cast} & 1 & 0.659 [0.576, 0.752] & 0.725 & 0.736 [0.673, 0.802] & 0.806 & 0.668 [0.599, 0.745] & 0.753 \\
 & 2 & 0.368 [0.288, 0.460] & 0.512 & 0.510 [0.439, 0.603] & 0.699 & 0.478 [0.409, 0.559] & 0.620 \\
 & 3 & 0.024 [0.008, 0.045] & 0.100 & 0.042 [0.020, 0.069] & 0.196 & 0.163 [0.113, 0.222] & 0.392 \\
\midrule
\cfd{inundation} & 1 & 0.703 [0.617, 0.787] & 0.759 & 0.804 [0.742, 0.868] & 0.846 & 0.748 [0.681, 0.819] & 0.789 \\
 & 2 & 0.585 [0.508, 0.674] & 0.674 & 0.688 [0.616, 0.759] & 0.787 & 0.635 [0.570, 0.705] & 0.737 \\
 & 3 & 0.326 [0.265, 0.395] & 0.390 & 0.472 [0.408, 0.545] & 0.607 & 0.398 [0.338, 0.463] & 0.505 \\
\midrule
\cfd{smoke\_haze} & 1 & 0.674 [0.582, 0.765] & 0.732 & 0.751 [0.675, 0.823] & 0.817 & 0.668 [0.601, 0.735] & 0.747 \\
 & 2 & 0.510 [0.428, 0.599] & 0.595 & 0.629 [0.567, 0.704] & 0.764 & 0.528 [0.447, 0.615] & 0.655 \\
 & 3 & 0.148 [0.104, 0.197] & 0.301 & 0.320 [0.268, 0.384] & 0.527 & 0.407 [0.332, 0.489] & 0.579 \\
\midrule
\cfd{fire\_warm\_tint} & 1 & 0.671 [0.581, 0.769] & 0.717 & 0.552 [0.468, 0.642] & 0.644 & 0.608 [0.531, 0.688] & 0.687 \\
 & 2 & 0.570 [0.472, 0.675] & 0.632 & 0.231 [0.167, 0.308] & 0.324 & 0.386 [0.313, 0.472] & 0.497 \\
 & 3 & 0.401 [0.313, 0.489] & 0.477 & 0.086 [0.055, 0.121] & 0.165 & 0.145 [0.103, 0.194] & 0.255 \\
\midrule
\cfd{rain\_streaks} & 1 & 0.650 [0.554, 0.753] & 0.728 & 0.674 [0.589, 0.759] & 0.759 & 0.611 [0.527, 0.701] & 0.735 \\
 & 2 & 0.220 [0.165, 0.288] & 0.353 & 0.217 [0.170, 0.268] & 0.497 & 0.341 [0.279, 0.410] & 0.474 \\
 & 3 & 0.003 [0.000, 0.010] & 0.023 & 0.003 [0.000, 0.010] & 0.037 & 0.056 [0.032, 0.085] & 0.143 \\
\midrule
\cfd{motion\_blur} & 1 & 0.015 [0.003, 0.030] & 0.029 & 0.003 [0.000, 0.010] & 0.012 & 0.107 [0.071, 0.152] & 0.149 \\
 & 2 & 0.000 [0.000, 0.000] & 0.005 & 0.000 [0.000, 0.000] & 0.003 & 0.018 [0.003, 0.036] & 0.011 \\
 & 3 & 0.000 [0.000, 0.000] & 0.000 & 0.000 [0.000, 0.000] & 0.000 & 0.003 [0.000, 0.011] & 0.002 \\
\midrule
\cfd{low\_light} & 1 & 0.451 [0.365, 0.540] & 0.537 & 0.000 [0.000, 0.000] & 0.001 & 0.047 [0.027, 0.073] & 0.064 \\
 & 2 & 0.110 [0.064, 0.163] & 0.176 & 0.000 [0.000, 0.000] & 0.000 & 0.000 [0.000, 0.000] & 0.000 \\
 & 3 & 0.000 [0.000, 0.000] & 0.010 & 0.000 [0.000, 0.000] & 0.000 & 0.000 [0.000, 0.000] & 0.000 \\
\midrule
\cfd{dust\_haze} & 1 & 0.599 [0.514, 0.686] & 0.665 & 0.638 [0.569, 0.711] & 0.758 & 0.570 [0.494, 0.650] & 0.662 \\
 & 2 & 0.344 [0.266, 0.423] & 0.469 & 0.318 [0.257, 0.386] & 0.504 & 0.315 [0.249, 0.384] & 0.478 \\
 & 3 & 0.098 [0.060, 0.139] & 0.208 & 0.071 [0.045, 0.102] & 0.220 & 0.104 [0.066, 0.151] & 0.263 \\
\bottomrule
\end{tabular}
\end{table}

\begin{table}[p]
\centering\scriptsize
\caption{Full results, SARD. Recall at the frozen operating point with bootstrap 95\% interval (1{,}000 image resamples), and threshold-independent mAP$_{50}$. Thresholds: Faster R-CNN 0.96, RT-DETR 0.60, YOLOv11 0.30.}
\label{tab:full_sard}
\setlength{\tabcolsep}{3pt}
\begin{tabular}{lccccccc}
\toprule
 & & \multicolumn{2}{c}{Faster R-CNN} & \multicolumn{2}{c}{RT-DETR} & \multicolumn{2}{c}{YOLOv11} \\
\cmidrule(lr){3-4}\cmidrule(lr){5-6}\cmidrule(lr){7-8}
Corruption & Sev. & Recall [95\% CI] & mAP$_{50}$ & Recall [95\% CI] & mAP$_{50}$ & Recall [95\% CI] & mAP$_{50}$ \\
\midrule
Clean & -- & 0.879 [0.858, 0.898] & 0.896 & 0.888 [0.869, 0.907] & 0.929 & 0.891 [0.870, 0.910] & 0.928 \\
\midrule
\cfd{water\_glare} & 1 & 0.842 [0.819, 0.865] & 0.859 & 0.851 [0.828, 0.873] & 0.888 & 0.859 [0.838, 0.880] & 0.883 \\
 & 2 & 0.768 [0.742, 0.795] & 0.773 & 0.789 [0.764, 0.818] & 0.813 & 0.809 [0.784, 0.834] & 0.836 \\
 & 3 & 0.629 [0.599, 0.661] & 0.609 & 0.646 [0.617, 0.674] & 0.653 & 0.666 [0.637, 0.696] & 0.686 \\
\midrule
\cfd{turbidity\_cast} & 1 & 0.799 [0.774, 0.822] & 0.846 & 0.791 [0.765, 0.816] & 0.891 & 0.796 [0.770, 0.820] & 0.883 \\
 & 2 & 0.538 [0.506, 0.569] & 0.656 & 0.476 [0.441, 0.509] & 0.699 & 0.575 [0.544, 0.606] & 0.756 \\
 & 3 & 0.180 [0.155, 0.208] & 0.264 & 0.074 [0.057, 0.091] & 0.144 & 0.191 [0.167, 0.214] & 0.408 \\
\midrule
\cfd{inundation} & 1 & 0.847 [0.825, 0.867] & 0.875 & 0.843 [0.821, 0.866] & 0.900 & 0.861 [0.838, 0.881] & 0.907 \\
 & 2 & 0.730 [0.701, 0.759] & 0.793 & 0.720 [0.692, 0.747] & 0.823 & 0.744 [0.717, 0.770] & 0.816 \\
 & 3 & 0.489 [0.455, 0.520] & 0.534 & 0.459 [0.426, 0.492] & 0.560 & 0.495 [0.464, 0.524] & 0.575 \\
\midrule
\cfd{smoke\_haze} & 1 & 0.811 [0.787, 0.835] & 0.856 & 0.779 [0.753, 0.808] & 0.870 & 0.810 [0.787, 0.833] & 0.882 \\
 & 2 & 0.625 [0.597, 0.655] & 0.714 & 0.615 [0.585, 0.645] & 0.763 & 0.667 [0.637, 0.697] & 0.787 \\
 & 3 & 0.303 [0.273, 0.332] & 0.402 & 0.319 [0.289, 0.350] & 0.486 & 0.391 [0.358, 0.422] & 0.573 \\
\midrule
\cfd{fire\_warm\_tint} & 1 & 0.799 [0.774, 0.821] & 0.835 & 0.724 [0.697, 0.751] & 0.854 & 0.688 [0.660, 0.716] & 0.825 \\
 & 2 & 0.604 [0.574, 0.632] & 0.687 & 0.368 [0.337, 0.398] & 0.562 & 0.310 [0.278, 0.341] & 0.500 \\
 & 3 & 0.407 [0.375, 0.439] & 0.507 & 0.118 [0.098, 0.136] & 0.251 & 0.099 [0.081, 0.120] & 0.222 \\
\midrule
\cfd{rain\_streaks} & 1 & 0.871 [0.851, 0.891] & 0.896 & 0.886 [0.867, 0.906] & 0.928 & 0.893 [0.875, 0.912] & 0.927 \\
 & 2 & 0.852 [0.831, 0.873] & 0.876 & 0.879 [0.859, 0.900] & 0.927 & 0.882 [0.863, 0.903] & 0.928 \\
 & 3 & 0.778 [0.754, 0.804] & 0.836 & 0.840 [0.818, 0.861] & 0.915 & 0.855 [0.833, 0.877] & 0.920 \\
\midrule
\cfd{motion\_blur} & 1 & 0.772 [0.747, 0.798] & 0.815 & 0.781 [0.755, 0.806] & 0.872 & 0.792 [0.765, 0.817] & 0.871 \\
 & 2 & 0.567 [0.535, 0.602] & 0.663 & 0.489 [0.457, 0.520] & 0.628 & 0.613 [0.581, 0.646] & 0.736 \\
 & 3 & 0.398 [0.365, 0.432] & 0.505 & 0.329 [0.297, 0.361] & 0.430 & 0.478 [0.443, 0.512] & 0.629 \\
\midrule
\cfd{low\_light} & 1 & 0.419 [0.389, 0.452] & 0.546 & 0.260 [0.233, 0.288] & 0.421 & 0.184 [0.157, 0.212] & 0.247 \\
 & 2 & 0.077 [0.060, 0.094] & 0.142 & 0.013 [0.007, 0.019] & 0.049 & 0.007 [0.003, 0.012] & 0.018 \\
 & 3 & 0.000 [0.000, 0.000] & 0.000 & 0.000 [0.000, 0.000] & 0.001 & 0.000 [0.000, 0.000] & 0.000 \\
\midrule
\cfd{dust\_haze} & 1 & 0.736 [0.707, 0.763] & 0.804 & 0.665 [0.634, 0.695] & 0.803 & 0.674 [0.645, 0.703] & 0.796 \\
 & 2 & 0.472 [0.439, 0.506] & 0.574 & 0.269 [0.238, 0.299] & 0.441 & 0.273 [0.242, 0.301] & 0.453 \\
 & 3 & 0.160 [0.136, 0.184] & 0.242 & 0.050 [0.037, 0.064] & 0.111 & 0.058 [0.044, 0.074] & 0.114 \\
\bottomrule
\end{tabular}
\end{table}

\end{document}